\documentclass[letterpaper, 10 pt, conference]{ieeeconf}  

\IEEEoverridecommandlockouts                              

\usepackage{graphics} 
\usepackage{epsfig} 
\usepackage{mathptmx} 
\usepackage{times} 
\usepackage{amsmath} 
\usepackage{amssymb}  
\usepackage{cite}
\usepackage{fancyhdr}  

\makeatletter
\let\NAT@parse\undefined
\makeatother
\usepackage{hyperref}

\title{\LARGE \bf
WS-3D-Lane: Weakly Supervised 3D Lane Detection With 2D Lane Labels
}

\author{Jianyong Ai$^{1}$ Wenbo Ding$^{1}$ Jiuhua Zhao$^{1}$ Jiachen Zhong$^{1}$
\thanks{$^{1}$Jianyong Ai, Wenbo Ding, Jiuhua Zhao and Jiachen Zhong are with SAIC AI Lab, 62 Weicheng Rd., Shanghai, China (Email:{\tt\small \{aijianyong, dingwenbo, zhaojiuhua, zhongjiachen\}@saicmotor.com})}%
}

\begin{document}

\maketitle
\thispagestyle{fancy}
\renewcommand\headrulewidth{0pt} 
\lfoot{}
\cfoot{\small{This work has been submitted to the IEEE for possible publication. Copyright may be transferred without notice, after which this version may no longer be accessible.}}
\rfoot{}
\pagestyle{empty}

\begin{abstract}

Compared to 2D lanes, real 3D lane data is difficult to collect accurately. In this paper, we propose a novel method for training 3D lanes with only 2D lane labels, called weakly supervised 3D lane detection \textsl{WS-3D-Lane}. By assumptions of constant lane width and equal height on adjacent lanes, we indirectly supervise 3D lane heights in the training. To overcome the problem of the dynamic change of the camera pitch during data collection, a camera pitch self-calibration method is proposed. In anchor representation, we propose a double-layer anchor with a improved non-maximum suppression (NMS) method, which enables the anchor-based method to predict two lane lines that are close. Experiments are conducted on the base of 3D-LaneNet under two supervision methods. Under weakly supervised setting, our \textsl{WS-3D-Lane} outperforms previous 3D-LaneNet: F-score rises to 92.3\% on Apollo 3D synthetic dataset, and F1 rises to 74.5\% on ONCE-3DLanes. Meanwhile, \textsl{WS-3D-Lane} in purely supervised setting makes more increments and outperforms state-of-the-art. To the best of our knowledge, \textsl{WS-3D-Lane} is the first try of 3D lane detection under weakly supervised setting.
\end{abstract}

\section{INTRODUCTION}

\begin{figure}[tbp]
    \centering
    \includegraphics[width=8cm]{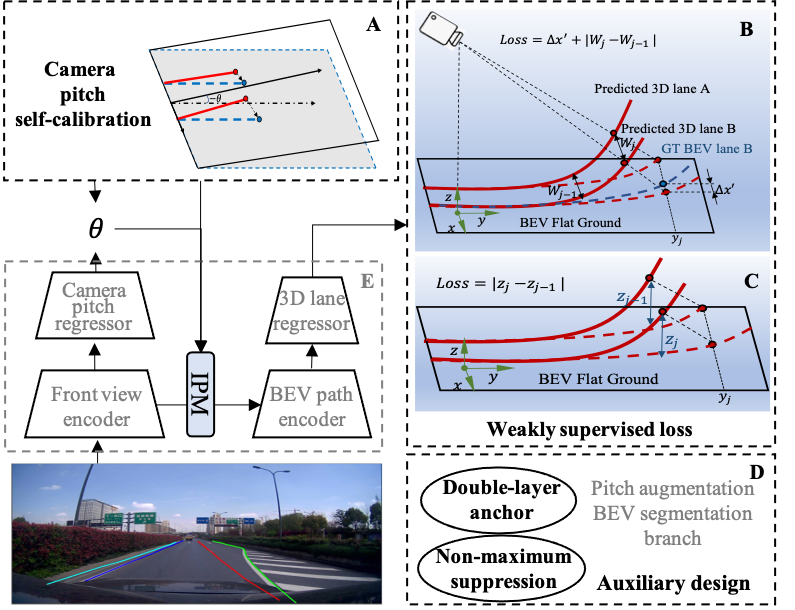}
    \caption{\textbf{Framework for \textsl{WS-3D-Lane}.} We use 3D-LaneNet\cite{20193Dlanenet} (E) as our basic network. Our weakly supervised loss function is designed in the assumptions of constant lane width (B) and equal height on adjacent lane lines (C). Camera pitch self-calibration method (A) is proposed for calculate the label of camera pitch per frame in real scenes. Auxiliary designs (D) are adopted in the training, including double-layer anchor, improved non-maximum suppression, BEV segmentation branch\cite{2022PersFormer} and camera pitch augmentation\cite{2022ReGenlanenet}. Bold text is our major contributed designs.}
    \label{fig:1}
\end{figure}

Image-based lane detection is an important perception task in autonomous driving. Traditional methods \cite{2021LaneAF, 2019LearningLL, 2021CondLane} detect or segment lanes in image domain and then project the results onto a flat ground, which is inaccurate in case of uphill or downhill and may lead to dangerous behavior of an autonomous vehicle. Compared to 2D lane, 3D lane detection\cite{2020Genlanenet,20193Dlanenet,2022PersFormer,2022ONCE3DLanes} can directly obtain the slope information of the lane, to help autonomous vehicles make better decisions. However, the data collection of 3D lanes in real-world is very difficult. Current published datasets, such as ONCE-3DLanes\cite{2022ONCE3DLanes, 2021OneMS} and OpenLane\cite{2022PersFormer}, are collected by LIDAR and front-view images with 2D labels. 3D points in LIDAR are converted to camera coordinate system and picked out by 2D lane labels. Tricks like filters and curve fitting are adopted to reduce errors. The quality of these datasets has room to improve. In the last decade, abundant high quality 2D lanes are collected \cite{2017CULane,2020CurveLane,2017vpgnet,2018ADF}. An intuitive thinking is fully taking advantage of these 2D lane data. Therefore, we propose a weakly supervised training method for 3D lane detection, called \textsl{WS-3D-Lane}, which uses 2D lane labels to supervise the training of 3D lane detection. The framework of \textsl{WS-3D-Lane} is shown in Fig.\ref{fig:1}. Based on the assumption of constant lane width and equal height on adjacent lane lines, our \textsl{WS-3D-Lane} utilizes the predicted lane width difference (Fig.\ref{fig:1}B) and height difference on adjacent lane lines (Fig.\ref{fig:1}C) to supervise the height of lane line. 

Besides lane label, camera extrinsic is also crucial in 3D lane detection, especially for camera pitch angle \cite{20193Dlanenet,2022CLGo}. But camera pitch often suffers from shake and cannot be dynamically calibrated during data collection. Previous data collection method used the same camera extrinsic for every frame\cite{2022PersFormer}. It might cause large errors on the ground truth in world coordinate system. Therefore, we proposed camera pitch self-calibration method (Fig.\ref{fig:1}A), to get the label of camera pitch per frame. 

Furthermore, we propose double-layer anchor with a improved non-maximum suppression (NMS) method which are able to improve the performance under both weakly supervised and purely supervised setting. Combined with two useful techniques: Picth augmentation\cite{2022ReGenlanenet} and BEV segmentation\cite{2022PersFormer}, our auxiliary designs (Fig.\ref{fig:1}D) help model achieve the state-of-the-art performance.

Our contributions are summarized as follows:
\begin{itemize}
    \item For the first time, our \textsl{WS-3D-Lane} provides a weakly supervised 3D lane detection method with only 2D lane labels, and outperforms previous 3D-LaneNet in the evaluation.  
    \item Our camera pitch self-calibration method provides a way to calculate accurate camera pitch per frame.
    \item With our auxiliary designs, our \textsl{WS-3D-Lane} is able to achieve the state-of-the-art level performance under both purely and weakly supervised setting.
\end{itemize}

\section{RELATED WORKS}

\subsection{3D Lane Detection}

The methods of 3D lane detection could be divided into LIDAR-based\cite{2018laneLiDAR,2010laneLiDAR,2014LIDAR,2008lidar}, monocular\cite{20203Dlanenet+,2021DualATTENTION, 2020SemiLocal3DL} and multi-sensor\cite{2018Multi-Sensor,2019lidar-camera,2021lidar-camera}. Our work focuses on the monocular 3D lane detection. 

The pioneering work 3D-LaneNet\cite{20193Dlanenet} predicts camera pitch, and converts the multi-scale image-view features to BEV features by the inverse perspective mapping(IPM) projection. It predicts 3D lanes on BEV path using anchors. Each anchor consists of lane probability and several reference points along y-axis. Each point contains the visibility and position of x-offset and z. 3Dlanenet+\cite{20203Dlanenet+} uses anchor-free representations to predict 3D lanes and gets better results on their private dataset. Gen-laneNet\cite{2020Genlanenet} makes a great contribution on the anchor representation. It predicts 2D points on BEV flat ground with the lane height, and converts to 3D points by geometric relationship. This improvement makes a great better performance on Apollo-Sim-3D. Persformer\cite{2022PersFormer} achieves the state-of-the-art on three datasets via replacing IPM with transformer and adding several auxiliary tasks. However, the previous anchor representation is not able to predict two close lane lines, such as forks and curb with nearby lane line. We propose the double-layer anchor to solve this problem. 
Reference \cite{2022ReGenlanenet} is the first literature to use lane width in loss function. It contributes an auxiliary loss function to keep the same change rate of lane width and uses a greedy matching algorithm to calculate lane width. We use the lane width differently. Firstly, the lane width is calculated by the equivalent short and straight lines. Secondly, we focus on the weakly supervised setting and the constant lane width assumption is the main supervised signal for lane height without LIDAR. 

\subsection{Weakly Supervised 3D Perception}

Previous works on 3D lane detection are fully supervised, but our work focuses on weakly supervised setting. Weakly supervised methods are widely used in 3D points segmentation and object detection \cite{2020WeaklyS3, 2020WeaklyS3L, 2021OneTO}, to save the cost of annotation of 3D point cloud. These methods are proposed to supervise the lacked part by rules and assumptions in different conditions, e.g. labeled 2D images with unlabeled 3D point cloud \cite{2021FGR} and labeled virtual scene with unlabeled real scene \cite{2022BackTR}. To the best of our knowledge, no previous literature focuses on the weakly supervised 3D lane detection, and our \textsl{WS-3D-Lane} is the first work to explore it. For the difficulty of collecting 3D lane labels in real-world, \textsl{WS-3D-Lane} provides a novel method using 2D lane labels to supervise the 3D lane detection.

\section{METHOD}

 For weakly supervised 3D lane detection, 2D lane labels should be converted to BEV flat ground with camera pitch and camera height first. Then we use the ground truth on BEV flat ground and our assumptions to supervise the model training. 
 Following previous literature \cite{2020Genlanenet, 2022ReGenlanenet}, the world coordinate center is the projection point of the camera center on the ground, and its axes are shown in Fig.\ref{fig:1}B. Camera roll and yaw are set to 0.

\subsection{Assumption of Constant 3D Lane Width}

\label{S3A}

\begin{figure}[htbp]
    \centering
    \includegraphics[width=8cm]{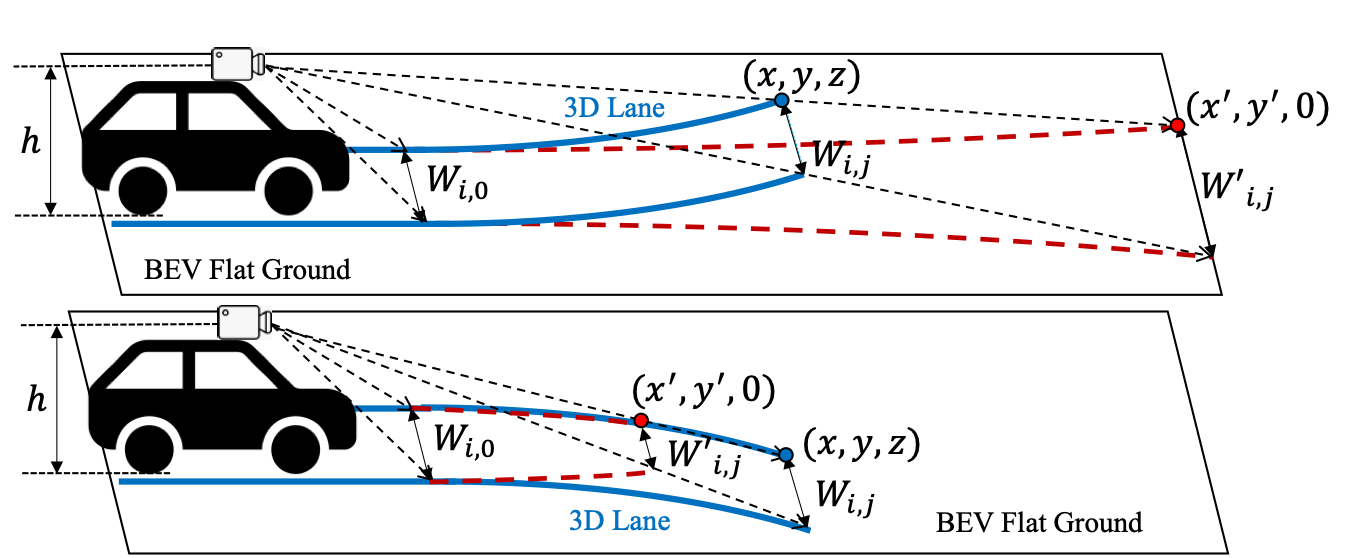}
    \caption{\textbf{Changes of lane width on BEV flat ground when 3D lane is not flat.}} \label{fig:2}
\end{figure}

The height of the 3D lane is implicitly included in the BEV panel. As shown in Fig. \ref{fig:2}, when the 3D lane is not flat, the lane width on the BEV flat ground changes. It tends to be wider when the front road is uphill and narrower when the front road is downhill. Equation (1) shows the geometric relationship between the point on the 3D lane $(x, y, z)$ and its corresponding 2D projection lane point on the BEV flat ground $(x', y', 0)$\cite{2020Genlanenet}, from which the relationship between the 3D lane width $W_{i,j}$ and the 2D lane width on the BEV flat ground $W'_{i,j}$ can be derived as in Equation (2), where $i\in\{0, 1, 2 ...\}$ means the serial number of lane line along the x-axis and $j\in\{0, 1, 2 ...\}$ means the serial number of anchor points along the y-axis. $W_{i,j}$ is the lane width between lane line $i$ and $i-1$ at reference point $j$. 

\begin{equation}
\frac{x}{x'} = \frac{y}{y'} = \frac{h-z}{h} 
\end{equation}
\begin{equation}
\frac{W_{i,j}}{W'_{i,j}} = \frac{h-z_{i,j}}{h}
\end{equation}

In most cases, the lane width is fixed for the same lane, so we can assume that the lane width is a constant value $W_{i,0}$. At the closest point, i.e. $z\approx 0$, the width of the closest lane on BEV flat ground $W'_{i,0}$ is equal to $W_{i,0}$, as shown in Fig. \ref{fig:2}.

For a point $P_{i,j}$ on a lane line, the key of lane width calculation is to find its corresponding closest point $P_{i,j}^{0}$ on its neighbored lane line. As shown in Fig. \ref{fig:3}, we use several short and straight lines to represent the curves, so the point $P_{i,j}^{0}$ is approximated to the perpendicular foot of $P_{i,j}$ on straight line $P_{i-1,j}P_{i-1,j-1}$. If the lane line is represented by the anchor, $P_{i-1,j}$ and $P_{i-1,j-1}$ could be two neighbored anchor points on a line, and then the lane width could be calculated as follows:

\begin{figure}[htbp]
    \centering
    \includegraphics[width=8cm]{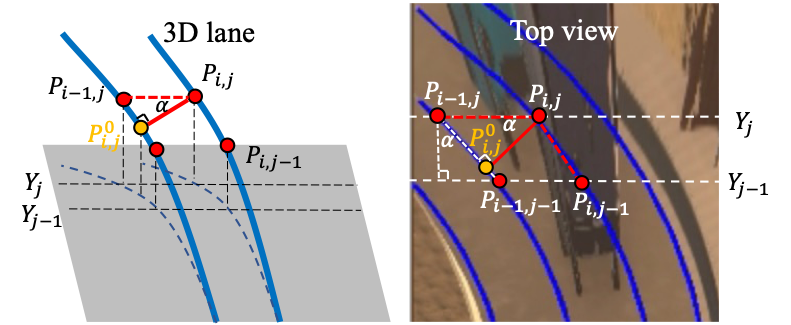}
    \caption{\textbf{3D lane width calculation on bend.}} \label{fig:3}
\end{figure}

\begin{equation}
\begin{aligned}
W_{i,j}&= |P_{i,j}P_{i,j}^{0}| \approx |P_{i,j}P_{i-1,j}|\cos\alpha \\
&= \frac{|P_{i,j}P_{i-1,j}||P_{i-1,j}P_{i-1,j-1}|_{x=0}}{|P_{i-1,j}P_{i-1,j-1}|}
\end{aligned}
\end{equation}
where, 
\begin{equation}
\begin{cases}|P_{i,j}P_{k,l}|= \left\|(x_{i,j} - x_{k,l}, y_{i,j} - y_{k,l}, z_{i,j} - z_{k,l})\right\|_2
 \\|P_{i,j}P_{k,l}|_{x=0} = \left\|(0, y_{i,j} - y_{k,l},z_{i,j} - z_{k,l})\right\|_2  
 \\x_{i,j} = \hat{x'}_{i,j}\frac{h-z_{i,j}}{h}
 \\y_{i,j} = \hat{y'}_{i,j}\frac{h-z_{i,j}}{h}
\end{cases}
\end{equation}
where, the symbol $\hat{x'}, \hat{y'}$ means the ground truth of $x', y'$.

Following previous literature \cite{2020Genlanenet}, for a predicted anchor $X_{i,j}^{A}$ and its corresponding ground truth $\hat{X}_{i,j}^{A} = \left \{ (\hat{p}_{i}^{A}, \hat{v}_{i,j}^{A}, \hat{x'}_{i,j}^{A}) \right \}$, each anchor point at a pre-defined position along y-axis $y_{j}$ predicts the x-offset $x'_{i,j}$ on BEV flat ground, the lane height $z_{i,j}$, and the visibility $v_{i,j}$. $\hat{p}_{i}^{A}$ is the probability of lane line in the anchor $i$. A loss function of weakly supervised 3D lane can be written as follows:
\begin{equation}
L_{width} = \sum_{i=1}^{N_p-1}\sum_{j=1}^{Y-1}\left\|W_{i,j} - W_{i,j-1}\right\|_{1}
\end{equation}
where, $N_p$ is the number of the anchors where $\hat{p}_{i}^{A} = 1$. $Y$ is the number of y steps in a anchor. This loss function is to force the model to predict lane with equal width.

\subsection{Assumption of Equal Height on Adjacent Lanes}
\label{S3B}
At the same distance along the y-axis, we assume the height of the lane line is equal to the height of its adjacent lanes. Therefore, in Fig. \ref{fig:3}, the height of point $P_{i,j}$ is equal to $P_{i-1,j}$, and a loss function of weakly supervised 3D lane can be written as follows: 
\begin{equation}
L_{height} = \sum_{i=1}^{N_p-1}\sum_{j=1}^{Y-1}\left\|z_{i,j} - z_{i-1,j}\right\|_{1}
\end{equation}

\subsection{Camera Pitch Self-calibration} 
\begin{figure}[htbp]
    \centering
    \includegraphics[width=6.5cm]{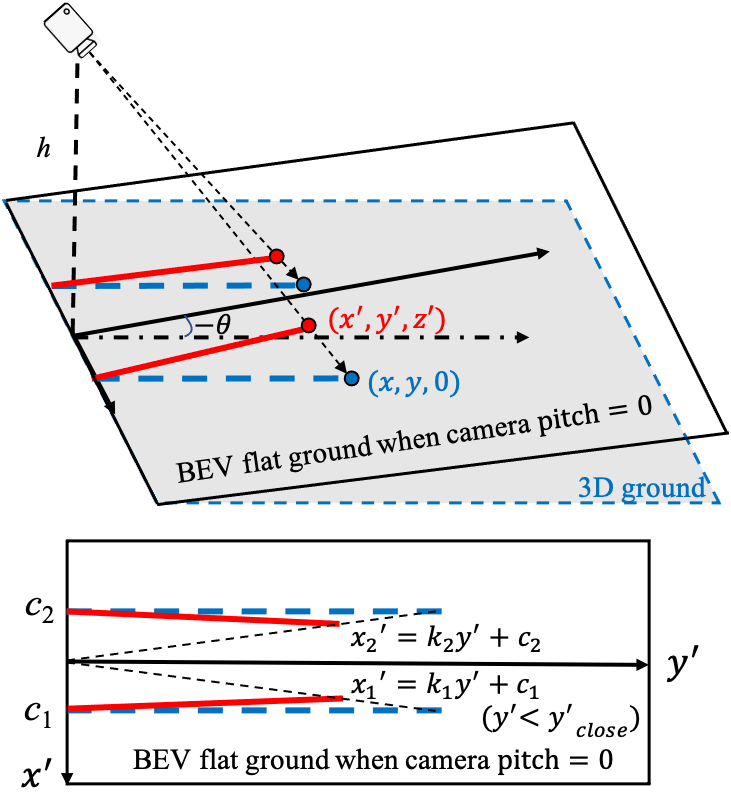}
    \caption{\textbf{Camera pitch self-calibration.}} \label{fig:4}
\end{figure}

The camera pitch self-calibration for each frame is necessary for transforming 2D labels from front-view to BEV flat ground\cite{2020Genlanenet}. In the self-calibration, we assume that the lane height is 0 when it is close enough, i.e. the $y'<y_{close}$. Therefore, when the lane lines are not parallel on BEV flat ground, the camera pitch should be corrected. As shown in Fig. \ref{fig:4}, we first assume camera pitch = 0 and transform the 2D lane lines to the BEV flat ground. Secondly, the transformed 2D lane lines are fitted by straight lines at $y'<y_{close}$. Therefore, the correct camera pitch $\theta$ is the angle between the 3D ground and the BEV flat ground under the assumption of camera pitch = 0. $\theta$ can be calculated as follows:
\begin{equation}
\begin{split}
\begin{cases} \frac{x'_{1}}{x_{1}}=\frac{x'_{2}}{x_{2}}=\frac{y'}{y}=\frac{h-z'}{h}
 \\x'_1=k_1y'_1+c_1
 \\x'_2=k_2y'_2+c_2
 \\-\tan \theta=\frac{z^{\prime}}{y^{\prime}}
\end{cases} \\
\Rightarrow \theta=\tan^{-1}\left(h\frac{k_{2}-k_{1}}{c_{2}-c_{1}}\right)
\end{split}
\end{equation}

\begin{figure}[htbp]
    \centering
    \includegraphics[width=6.5cm]{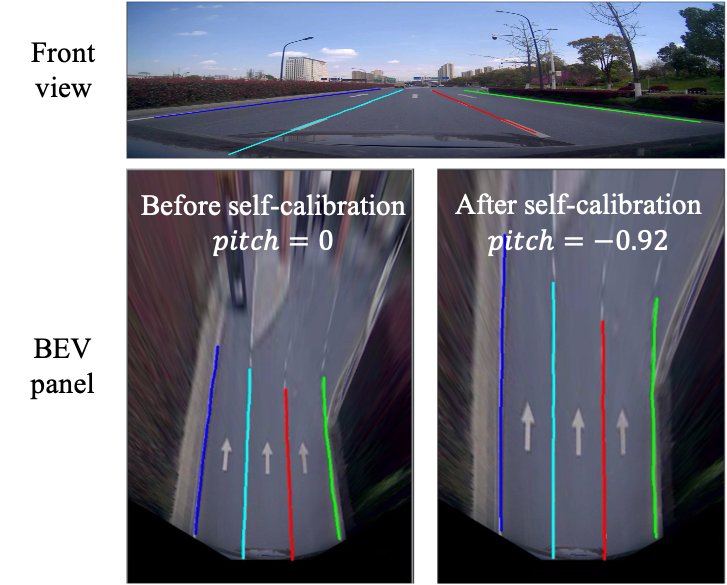}
    \caption{\textbf{Camera pitch self-calibration result.}} \label{fig:5}
\end{figure}

A calibration result is shown in Fig. \ref{fig:5}. The front view image shows the automobile driving on flat ground. Before self-calibration, the lanes have a toe-in angle. After the calibration, the lane lines become parallel and the ground looks flat. On Apollo-Sim-3D, the average error between the ground truth and the calibrated camera pitch is $0.11^{\circ}$. When the camera pitch is predicted in the network, L1 loss is used to regress the camera pitch as $L_{pitch}$. 

\subsection{Auxiliary Designs} 

\begin{figure}[htbp]
    \centering
    \includegraphics[width=8cm]{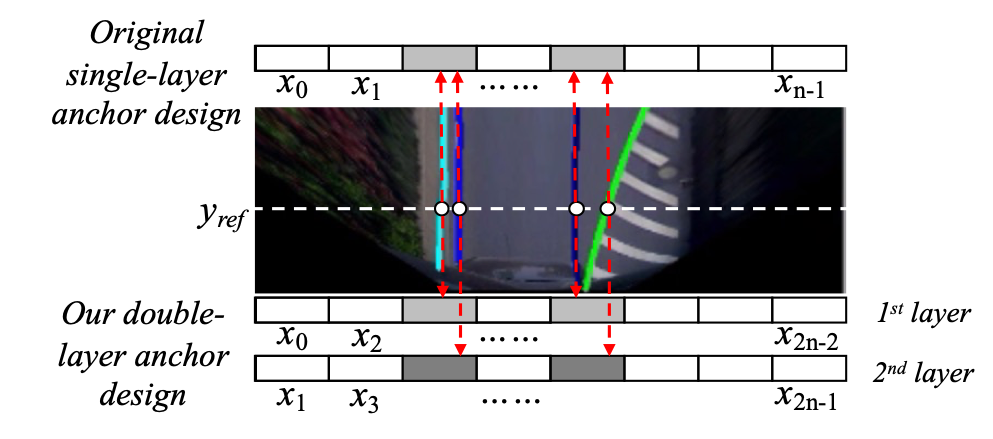}
    \caption{\textbf{Single-layer anchor design \cite{20193Dlanenet} v.s. Our double-layer anchor design.}} \label{fig:6}
\end{figure}

\textbf{Double-layer anchor}. Based on the work of 3D-LaneNet, we design the double-layer anchors to improve the accuracy in the case where the lane lines are too close, as shown in Fig. \ref{fig:6}. The first layer follows the 3D-LaneNet design\cite{20193Dlanenet}, and the second layer has the same x steps as the first layer. When there are two close lanes in the anchor, $\hat{p}_{i}^{A}$ of the second layer will be set to 1. The anchors of the first layer predict the left line of the two close lane lines, while the second layer anchors predict the right. However, the constant width assumption in the weakly supervised setting is usually invalid if two lines are too close to each other, e.g. fork road or curb with a nearby lane line as in Fig.\ref{fig:8}. Therefore, in real implementation, we simply drop $L_{width}$ and $L_{height}$ loss terms described in section \ref{S3A} and section \ref{S3B} when the second-layer anchor is activated during model training. 

\textbf{Non-maximum suppression}. To avoid multiple anchors predicting the same lane line, we provide an improved non-maximum suppression (NMS) method. For each anchor prediction $X_i^A$, the mean value of $x_i^A$ distance between $X_i^A$ and another neighbored anchor $X_k^A$ is calculated as $\bar{d_{i,k}}$. If $\bar{d_{i,k}} < d_{thresh}$ and $p_i^A<p_k^A$, $p_i^A$ will be set to 0.

Besides, we also employ pitch augmentation\cite{2022ReGenlanenet} and BEV segmentation branch\cite{2022PersFormer} in order to further improve our model performance. 

\subsection{Loss Function} 
The loss function of our \textsl{WS-3D-Lane} $L_{ws}$ consists of a purely supervised part on BEV flat ground $L_{bev}$, a weakly supervised part $L_{width} + L_{height}$, a BEV segmentation loss $L_{seg}$ and a camera pitch regression loss $L_{pitch}$.
\begin{equation}
L_{ws} = L_{bev} + L_{width} + L_{height} + L_{seg} + L_{pitch}
\end{equation}

The purely supervised part $L_{bev}$ of the loss function is the same with Gen-LaneNet\cite{2020Genlanenet} except the difference between prediction and label along z-axis $L_{z}$:
\begin{equation}
\begin{aligned}
L_{bev} &=-\sum_{i=0}^{N-1}\left(\hat{p}_{i} \log p_{i}+\left(1-\hat{p}_{i}\right)\left(\log \left(1-p_{i}\right)\right)\right.\\
&+\sum_{i=0}^{N-1}\sum_{j=0}^{Y-1} \hat{p}_{i} \cdot\left(\left\|\hat{v}_{i,j} \cdot\left(x'_{i,j}-\hat{x'}_{i,j}\right)\right\|_{1}\right) \\
&+\sum_{i=0}^{N-1}\sum_{j=0}^{Y-1} \hat{p}_{i} \cdot\left\|v_{i,j}-\hat{v}_{i,j}\right\|_{1}
\end{aligned}
\end{equation}
where $N$ is the number of anchors along x-axis. For comparison, the loss function of our \textsl{WS-3D-Lane} in purely supervised setting $L_{sup}$ use $L_{z}$ to train the lane height.
\begin{equation}
L_{sup} = L_{bev} + L_{z} + L_{seg} + L_{pitch}
\end{equation}
\begin{equation}
L_{z} = \sum_{i=0}^{N-1}\sum_{j=0}^{Y-1} \hat{p}_{i} \cdot\left(\left\|\hat{v}_{i,j} \cdot\left(z_{i,j}-\hat{z}_{i,j}\right)\right\|_{1}\right)
\end{equation}

\section{Experiments}

\begin{table*}[htbp]
\caption{Evaluation results on Apollo-Sim-3D}
\label{table_1}

\begin{center}
\begin{tabular}{l|cccccc}
\hline         \multicolumn{1}{c|}{Method}                          & F-Score(\%)$\uparrow$  & AP(\%)$\uparrow$ & x error near (m) $\downarrow$   & x error far (m)$\downarrow$  & z error near (m)$\downarrow$  & z error far (m)$\downarrow$  \\
\hline 3D-LaneNet\cite{20193Dlanenet}                       & 86.4    & 89.3 & 0.068            & 0.477           & 0.015            & 0.202           \\
Gen-lanenet\cite{2020Genlanenet}              &  88.1  & 90.1 & 0.061 
& 0.496  & 0.012 &0.214\\
3D-LaneNet\#\cite{2020Genlanenet}                 & 90.0    & 92.0 & -            & -           & -            & -           \\
3D-LaneNet*                 & 89.8    & 91.9 & 0.054            & 0.408           & 0.010            & 0.243           \\
Gen-lanenet-reconstruct\cite{2022ReGenlanenet}          & 91.9 &93.8 &0.049 &0.387 &0.008 &\textbf{0.213}
       \\
Persformer\cite{2022PersFormer}           & 92.9    & - & 0.054            & 0.356           & 0.010            & 0.234           \\
\textbf{WS-3D-Lane(ours)} & 92.3        & 94.6      & 0.060                 & 0.373                & 0.023                 & 0.233  \\
\textbf{WS-3D-Lane\_sup(ours)}    & \textbf{93.5}    & \textbf{95.7} & \textbf{0.027}                  &  \textbf{0.321}               & \textbf{0.006}                 &  0.215               
\end{tabular}
\end{center}
'near' means $y<40 m$ and 'far' means $y>=40 m$.
3D-LaneNet\# is 3D-LaneNet with anchor representation in Gen-Lanenet\cite{2020Genlanenet}.
\ 3D-LaneNet* is our reproduced 3D-LaneNet\#.
\ WS-3D-Lane is our weakly supervised 3D-Lane network.
\ WS-3D-Lane\_sup is WS-3D-Lane under purely supervised setting.

\end{table*}

\begin{figure*}[htbp]
    \centering
    \includegraphics[width=16cm]{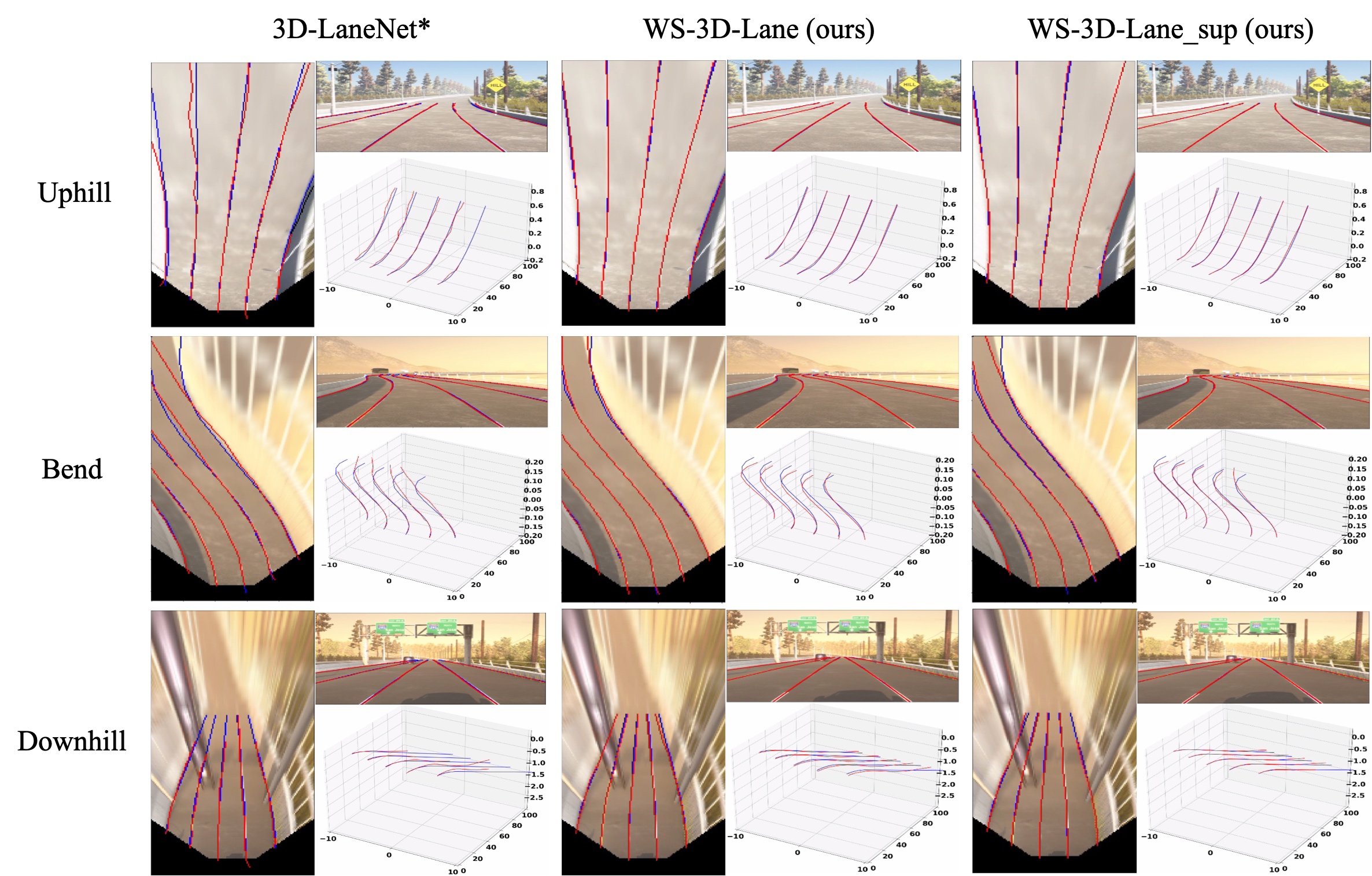}
    \caption{\textbf{Qualitative comparison on Apollo-Sim-3D.} Blue lines: ground truth, Red lines: prediction. Each visualized result consists of the front view image (right top), BEV panel (left), and 3D lanes (right down).} \label{fig:7}
\end{figure*}

\subsection{Experiments Setup}

\textbf{Datasets.} We conduct the experiments on both Apollo-Sim-3D and ONCE-3DLanes. Apollo-Sim-3D is a published synthetic 3D lane dataset\cite{2020Genlanenet} and is widely used\cite{2020Genlanenet,2022PersFormer,2022ReGenlanenet}. It provides 10.5K images with 3D lane labels and corresponding camera pitch. ONCE-3DLanes collect 211K images in real-world by LIDAR and auto-labeled 2D images, but the camera extrinsics is not recorded. 

\textbf{Comparison.} For a fair comparison with previous works, the camera extrinsics is assumed to be known on Apollo-Sim-3D and the evaluation method is following the previous literature\cite{2020Genlanenet}. On ONCE-3DLanes, the camera pitch is calculated per frame by our self-calibration method with $y_{close}=10m$ and the predicted lane lines are turned to the camera coordinate system to follow the same evaluation in \cite{2022ONCE3DLanes}. To show the upper limit of our work, \textsl{WS-3D-Lane} in the purely supervised setting, called \textsl{WS-3D-Lane\_sup}, is also compared in the experiment. The implementation details of \textsl{WS-3D-Lane\_sup} is the same with \textsl{WS-3D-Lane} except the loss function. \textsl{WS-3D-Lane} use equation (8) as loss function while \textsl{WS-3D-Lane\_sup} use equation (10).

\textbf{Implementation details.} Our experiments are carried out on our proposed \textsl{WS-3D-Lane} benchmark. On the two datasets, the input shape is $360 \times 480$, and the IPM feature map shape is $208 \times 128$. For model training, we use the Adam optimizer \cite{2015Adam} with the initial learning rate of $1e-3$ and the weight decay of $2e-4$. The learning rate is linearly annealed to $1e-7$ during training. We train the network with the batch size of 16 on one Nvidia Tesla V100 GPU. The training epochs are set to 100 on Apollo-Sim-3D and 30 on ONCE-3DLanes. For data augmentation, we adopt the random camera pitch noise with range $[-1^{\circ}, 1^{\circ} ]$. For anchor representation, we adopt the BEV region with x-range $[-10,10] m$ and use $Y_{ref}=5m$ to associate each anchor with its closet lane. The y-range is set to $[0,100] m$ on Apollo-Sim-3D and $[0, 50] m$ on ONCE-3DLanes. We design the y-reference points as \{0, 2.5, 5, 7.5, 10, 12.5, 15, 17.5, 20, 25, 30, 35, 40, 45, 50, 60, 70, 80, 90, 100\} $m$ on Apollo-Sim-3D and $1m$ per step on ONCE-3DLanes. In the post process, we adopt our improved NMS method with $d_{thresh}=0.05 m$ . 

\subsection{Results and Qualitative Comparison}

\textbf{Results.} The evaluation results on Apollo-Sim-3D are shown in Table \ref{table_1}. Compared to 3D-LaneNet\cite{20193Dlanenet}, 3D-LaneNet\#\cite{2020Genlanenet}, and our reproduction 3D-LaneNet*, our \textsl{WS-3D-Lane} consistently shows much better F-score and AP using only weakly supervised setting. Under purely supervised setting, our \textsl{WS-3D-Lane\_sup} performs 93.5\% on F-score outperforming previous state-of-the-art PersFormer's \cite{2022PersFormer} 92.9\%, and achieves state-of-the-art results on other metrics compared to previous works. As some scenes are inconsistent with our assumptions, \textsl{WS-3D-Lane} underperforms \textsl{WS-3D-Lane\_sup}, especially on z error. In Table \ref{table_2}, we report experimental results on ONCE-3DLanes. Under only weakly supervised setting, our \textsl{WS-3D-Lane} is able to archive similar results to previous supervised state-of-the-art method PersFormer\cite{2022PersFormer}. By using supervised setting, our \textsl{WS-3D-Lane\_sup} outperforms PersFormer\cite{2022PersFormer} on all metrics by a large margin.

\begin{table*}[htbp]
\caption{Evaluation results on ONCE-3DLanes}
\label{table_2}
\begin{center}
\begin{tabular}{l|cccc}
\hline        \multicolumn{1}{c|}{Method}                        & F1 (\%)$\uparrow$ & Precision (\%)$\uparrow$ & Recall (\%)$\uparrow$ & CD error (m)$\downarrow$ \\
\hline
3D-LaneNet\cite{2022ONCE3DLanes}                    & 44.73   & 61.46          & 35.16       & 0.127        \\
SALAD\cite{2022ONCE3DLanes}                     & 64.07   & 75.90          & 55.42       & 0.098        \\
Persformer\cite{2022PersFormer}                & 74.33   & 80.30          & 69.18       & 0.074        \\
\textbf{WS-3D-Lane(ours)} & 74.56   & 81.31          & 68.85       & 0.072    \\
\textbf{WS-3D-Lane\_sup(ours)}    & \textbf{77.02}    & \textbf{84.51}          & \textbf{70.75}       & \textbf{0.058} 
\end{tabular}
\end{center}
CD error: Chamfer distance between predicted lane line and ground truth. 
\ WS-3D-Lane is our weakly supervised 3D-Lane network.
\ WS-3D-Lane\_sup is our WS-3D-Lane under purely supervised setting.
\end{table*}

\begin{figure*}[htbp]
    \centering
    \includegraphics[width=16cm]{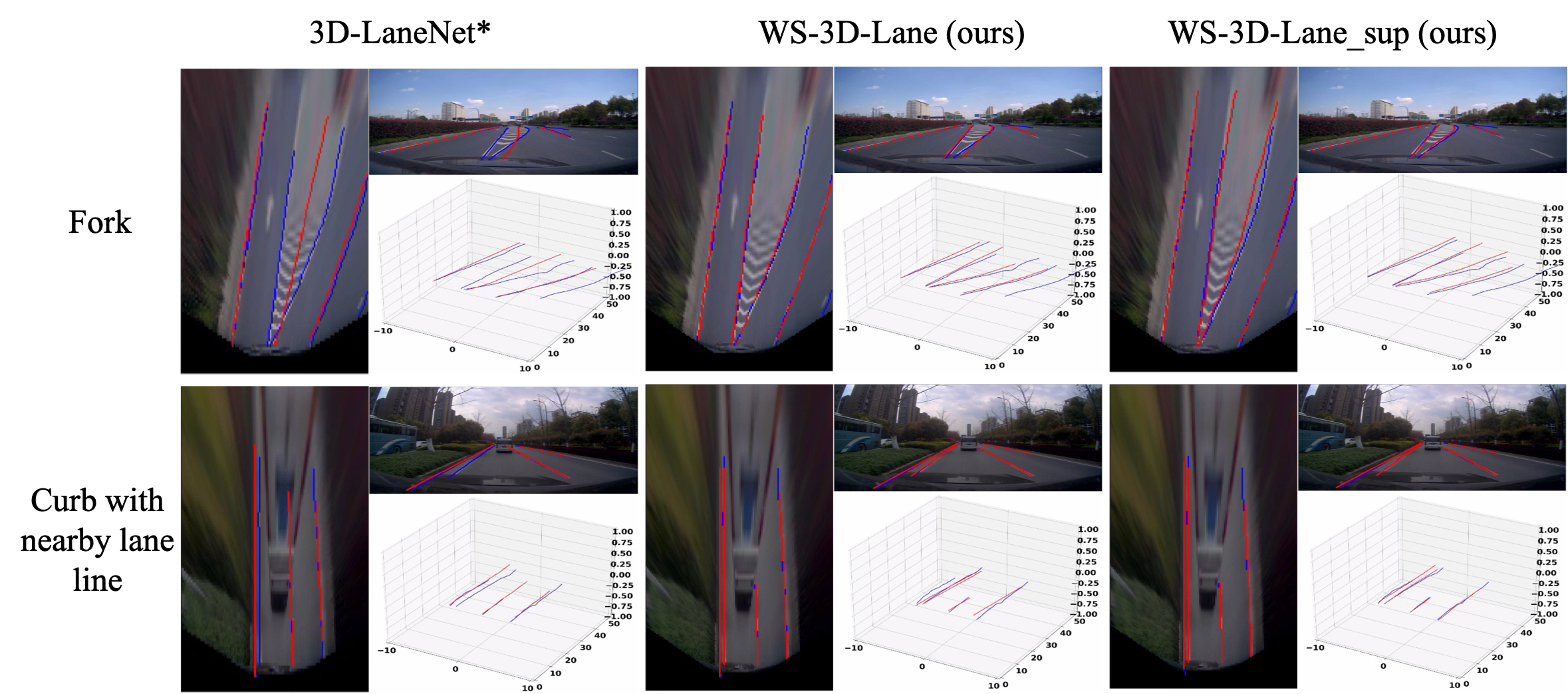}
    \caption{\textbf{Qualitative comparison on ONCE-3DLanes.} Blue lines: ground truth, Red lines: prediction.} 
    \label{fig:8}
\end{figure*}

\textbf{Qualitative comparison.} Figure \ref{fig:7} shows the qualitative comparison of reproduced 3D-LaneNet*, our \textsl{WS-3D-Lane} and our \textsl{WS-3D-Lane\_sup} on Apollo-Sim-3D. The predicted 3D lanes are projected onto the front view image and BEV panel. In the scenes of uphill and bend, 3D-LaneNet* predicts incorrect location along x-axis on far lane lines, but \textsl{WS-3D-Lane} and \textsl{WS-3D-Lane\_sup} predict it more correctly. In the scene of downhill, the predicted lane lines of 3D-LaneNet* are much shorter than the ground truth. It is caused by the prediction error in the visibility. \textsl{WS-3D-Lane} and \textsl{WS-3D-Lane\_sup} perform much better on this case with correct visibility prediction. Figure \ref{fig:8} shows the qualitative comparison of ONCE-3DLanes. In the scenes of fork and curb with nearby lane lines, 3D-LaneNet* is not able to predict two close lane lines at the same time, and the model tends to predict the center line of the two lane lines. After using the double-layer anchor, our \textsl{WS-3D-Lane} predicts two lane lines correctly. However, compared to \textsl{WS-3D-Lane\_sup}, \textsl{WS-3D-Lane} get a little larger errors on the height in all the scenes, which is reasonable because of the lack of direct supervision on lane height.

\subsection{Ablation Study}

We conduct the ablation study on ONCE-3DLanes to evaluate the impact of our designs and the results are shown in Table \ref{table_3}. As a result, high quality pitch angles are critical to 3D lane detection. With our per-frame self-calibrated pitch angle, our baseline model significantly outperforms 3D-LaneNet\cite{2022ONCE3DLanes} which uses only fixed camera extrinsics provided by ONCE\cite{2021OneMS}. Pitch augmentation contributes extra performance boost by augmenting the diversity of pitch angle. Double-layer anchor also improves the metric by equipping model with the ability to predict two close lane lines and the performance gain becomes even larger after combining it with our improved NMS method. BEV segmentation branch shows marginally improvements in supervised setting but brings side effects under weakly supervised setting, which might be caused by the errors of auto-labeled 2D lanes in the dataset.

\begin{table}[htbp]
\caption{Results of ablation study on ONCE-3DLanes}
\label{table_3}
\begin{center}
\begin{tabular}{l|cc|cc}
\hline                                
& \multicolumn{2}{c}{Supervised}  & \multicolumn{2}{c}{Weakly Supervised} \\
\hline \multicolumn{1}{c|}{Method}        & F1 $\uparrow$        & CD error $\downarrow$   & F1 $\uparrow$            & CD error $\downarrow$      \\
        \hline  
        3D-LaneNet\cite{2022ONCE3DLanes} &44.73 &0.127 &- &-\\
        \hline
        Baseline       & 70.07          & 0.059          & 68.73                  & 0.074                  \\
        +PA           & 74.27          & 0.059          & 72.23             & 0.072             \\
        +PA+DA      & 75.27          & \textbf{0.056}          & 73.83                 & \textbf{0.071}                  \\
        +PA+DA+NMS      & 76.13          & 0.057         &  \textbf{74.84}                 & \textbf{0.071}                  \\
        +PA+DA+NMS+BS & \textbf{77.02} & 0.058 & 74.56    & 0.072   
\end{tabular}
\end{center}
Baseline: 3D-laneNet\#\cite{2020Genlanenet} with our self-calibrated camera pitch and training hyperparameters. PA: pitch augmentation~\cite{2022ReGenlanenet}. DA: double-layer anchor. NMS: our improved NMS method. BS: BEV segmentation branch\cite{2022PersFormer}.
\end{table}

\section{CONCLUSIONS}
In this paper, we present a weakly supervised 3D lane detection method that allows the model training with only 2D labels by using the assumptions of constant lane width and equal height on adjacent lane lines. A self-calibration method is proposed to improve camera pitch quality of data. Several auxiliary designs are applied  in order to improve the overall model performance. We conduct comprehensive experiments to validate the effectiveness of our method. We believe a weakly supervised 3D lane detection approach by using only 2D annotations is valuable in both academic research and real industrial production, and we hope our \textsl{WS-3D-Lane} can stimulate more related research in the future.

\addtolength{\textheight}{-9cm}   




{
\bibliographystyle{ieeetr}
\bibliography{egbib}
}

\end{document}